\documentclass[conference]{IEEEtran}
\IEEEoverridecommandlockouts
\usepackage{cite}
\usepackage{amsmath,amssymb,amsfonts}
\usepackage{algorithmic}
\usepackage{graphicx}
\usepackage{textcomp}
\usepackage{subfig}
\usepackage{xcolor}
\def\BibTeX{{\rm B\kern-.05em{\sc i\kern-.025em b}\kern-.08em
    T\kern-.1667em\lower.7ex\hbox{E}\kern-.125emX}}
\begin{document}

%\title{Comparing AI approaches for Tetris Link}
\title{A New Challenge: Approaching Tetris Link with AI}

\author{\IEEEauthorblockN{Matthias Müller-Brockhausen, Mike Preuss, Aske Plaat}
\IEEEauthorblockA{\textit{Leiden Institute of Advanced Computer Science} \\
\textit{Leiden University}\\
The Netherlands \\
cog2020@mmb2.click}
}

\maketitle
% orient here: http://ieee-cog.org/2019/papers/paper_177.pdf
\begin{abstract}
Decades of research have been invested in making computer programs for playing games such as Chess and Go.
%AI for well-known games such as Chess and Go has experienced decades of research focusing on playing just a little better.
This paper focuses on a new game, %We will look into 
Tetris Link, a board game that is still lacking any scientific analysis.
Tetris Link has a large branching factor, hampering a traditional heuristic planning approach. %Besides assessing the branching factor (162) and first player advantage (not significant), we compare different approaches to automatically master the game, namely:
We explore heuristic planning and two other approaches: Reinforcement Learning, Monte Carlo tree search. We document our approach and report on their relative performance in a tournament. Curiously, the heuristic approach is stronger than the planning/learning approaches. However, experienced human players easily win the majority of the matches against the heuristic planning AIs. We, therefore, surmise that Tetris Link is more difficult than expected. We offer our findings to the community as a challenge to improve upon.
\end{abstract}


\section{Introduction}
Board games are favorite among AI researchers for experiments with intelligent decision making, building, for example, on works that analyze the game of Chess date back centuries\cite{philidor1790analysis,sprague1889different}.
Already in 1826, papers were published on machines that supposedly played Chess automatically\cite{bradford1826history}, although it was unclear whether the machine was still operated somehow by humans.
Nowadays, for some games, such as Chess~\cite{hsu1997deep} and Go~\cite{silver2016mastering,silver2017mastering}, we know for sure that there are algorithms that can, without the help of humans, automatically decide on a move, and even out-play the best human players. %In the game of Go, these algorithms even beat the world champion\cite{silver2016mastering,borowiec2016alphago}.
In this paper, we want to investigate a new game, Tetris Link, that has not yet received attention from researchers before, to the best of our knowledge %game and automated play approaches for the board game Tetris Link, because it has not been done yet 
(see section \ref{sec:relatedwork}). Tetris Link is a manual, multi-player version of the well-known video-game Tetris. It is played on a vertical "board", not unlike Connect-4. The game has a large branching factor, and since it is not immediately obvious how a strong computer program should be designed, we put ourselves to this task in this paper.
For that, we implement a digital version of the board game %(section \ref{sec:impl}) 
and take a brief look at the game's theoretic aspects (section \ref{sec:tetrislink}). Based on that theory, we develop heuristics for a minimax-based program that we also test against human players (section \ref{sec:heuristic}). Performance is limited, and we try other common AI approaches: Deep Reinforcement Learning (RL)~\cite{sutton2018introduction} and Monte Carlo tree search (MCTS)~\cite{browne2012survey}, approaches that were combined by Silver et al.\ in AlphaGo\cite{silver2017mastering}.
In section \ref{sec:exprl}, we look at experimental results RL, and in section \ref{sec:expmcts}, we look at MCTS as options to implement agents.
In our design of the game environment for the RL agent, we assess the impact of choices such as the reward on training success.
% We also briefly touch the topic of reproducibility %(section \ref{sec:exprepro}) in the deep learning field.
Finally, we compare the performance of these agents after letting them compete against each other in a tournament (section \ref{sec:exptournament}). To our surprise, the humans are stronger.

The main contribution of this paper is that we present to the community the challenge of implementing a well-playing computer program for Tetris Link.
This challenge is much harder than expected, and we provide evidence on why this might be the case, even for  the deterministic 2-player version (without dice) version of the game. The real Tetris Link can be played with four players using dice, which will presumably be even harder for an AI. 

We document our approach, implementing three players based on the three main AI game-playing approaches of heuristic planning, Monte Carlo Tree Search, and Deep Reinforcement Learning. To our surprise and regret, all players were handily beaten by human players. We respectfully offer our approach, code and experience to the community to improve upon. 

\section{Related Work}
\label{sec:relatedwork}
%Searching Google Scholar for the exact quoted term "Tetris Link" brings up three papers in total\cite{emptyscholar}. Two of them are entirely unrelated and talk about optical refrigerators, where there seems to be a phenomenon that is called "Tetris", which they "link" to a specific behaviour.
%The third paper
Few papers on Tetris Link exist in the literature. A single paper describes an experiment using Tetris Link~\cite{orensteindoes}. This work is about teaching undergraduates "business decisions" using the game Tetris Link.
To provide background on the game, we analyze the game in more depth in section \ref{sec:tetrislink}.

The AI approaches that we try have been successfully applied to a variety of board games~\cite{plaat2020learning}. Heuristic planning has been the standard approach in many games such as Othello, Checkers, and Chess~\cite{russell2016,schaeffer2002,hsu1997deep,van2002search,plaat1996}, MCTS has been used in a variety of applications such as Go, and GGP~\cite{coulom2006,browne2012survey,ruijl2013combining} and Deep RL has seen great success in Backgammon and Go~\cite{sutton2018introduction,van2008application,silver2016mastering,tesauro1989neurogammon}. Multi-agent MCTS has been presented in~\cite{galvan2014multi}.

\section{Tetris Link}
\label{sec:tetrislink}
\begin{figure}
\centering
\includegraphics[width=0.4\textwidth]{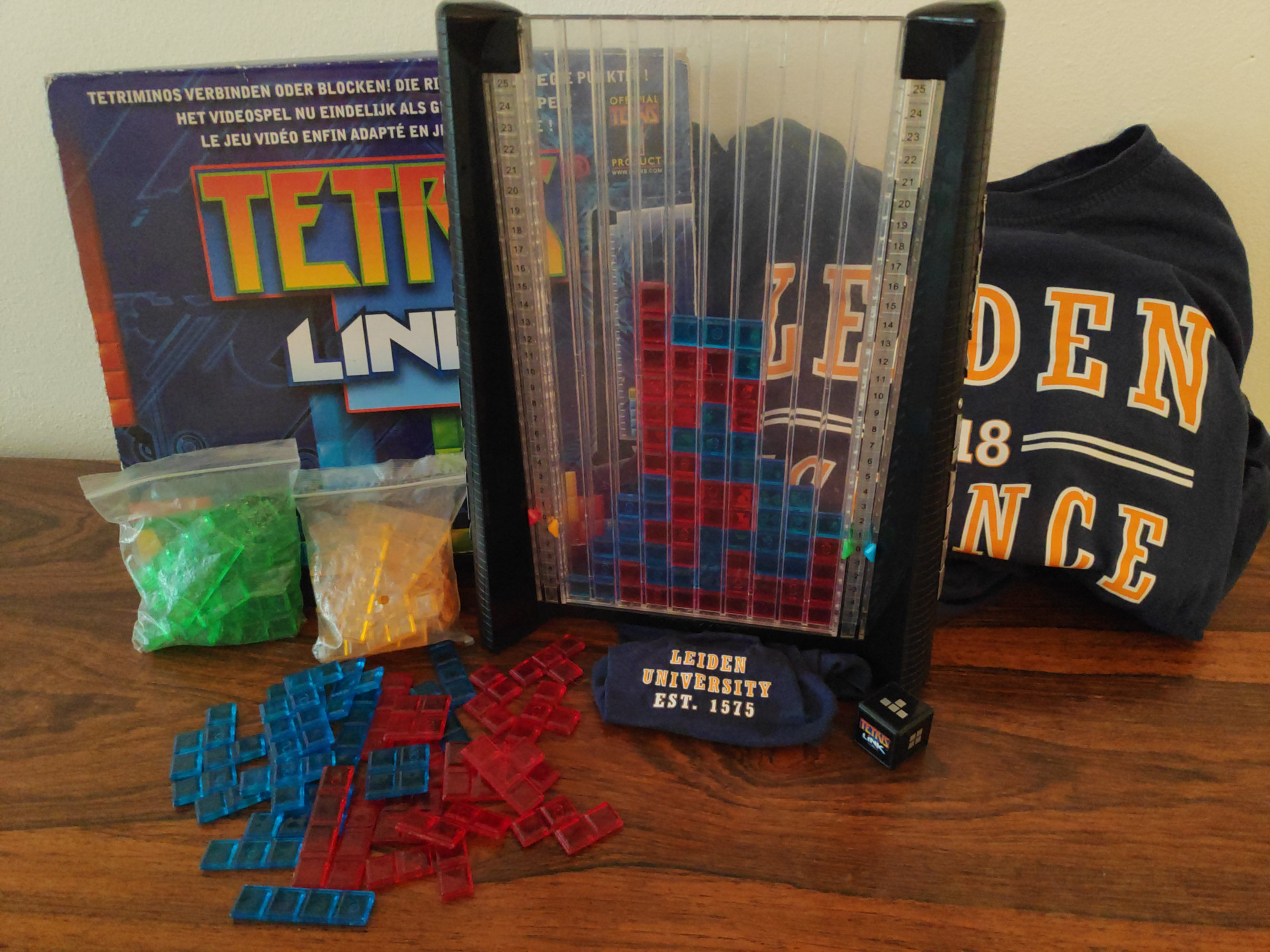}
\caption{A photo of the original Tetris Link board game. The colored indicators on the side of the board help to keep track of the score.}
\label{fig:original_game}
\end{figure}

Tetris Link, depicted in Figure \ref{fig:original_game}, is a turn-based board game for two to four players.
Just as the original Tetris video game, Tetris Link features a ten by twenty grid in which shapes called tetrominoes\footnote{A shape built from squares that touch each other edge-to-edge is called a polyomino\cite{demaine2007jigsaw}. Because they are made out of precisely four squares, these shapes are called tetromino\cite{wikitetromino}.} are placed on a board. This paper will refer to tetrominoes as blocks for brevity.
The five available block shapes are referred to as: \emph{I}, \emph{O}, \emph{T}, \emph{S}, \emph{L}\footnote{The \emph{S} and \emph{L} blocks may also be referred to as \emph{Z}\cite{burgiel1997lose} and \emph{J}\cite{carr2005applying}.}. Every shape has a small white dot, also in the original physical board game variant, to make it easier to distinguish individual pieces from each other.
Every player is assigned a colour for distinction and gets twenty-five blocks: five of each shape. In every turn, a player must place precisely one block. If no fitting blocks are available any more, then the player will be skipped. A player can never voluntarily skip if one of the available blocks fits somewhere in the board even if placing it is disadvantageous.
The game ends when no block of any player fits into the board any more.

\begin{figure}[t] 
    \centering
  \subfloat[No points.\label{fig:nopoints}]{%
       \includegraphics[width=0.3\linewidth]{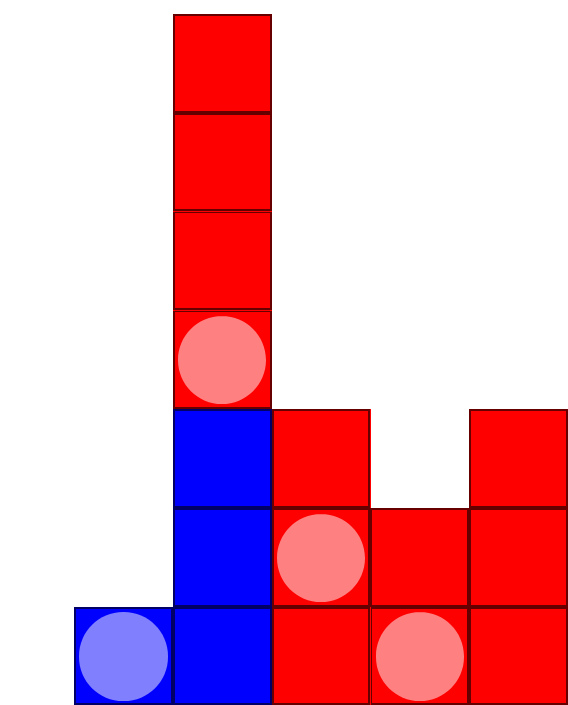}}
    \hfill
  \subfloat[Three points.\label{fig:groupthree}]{%
        \includegraphics[width=0.3\linewidth]{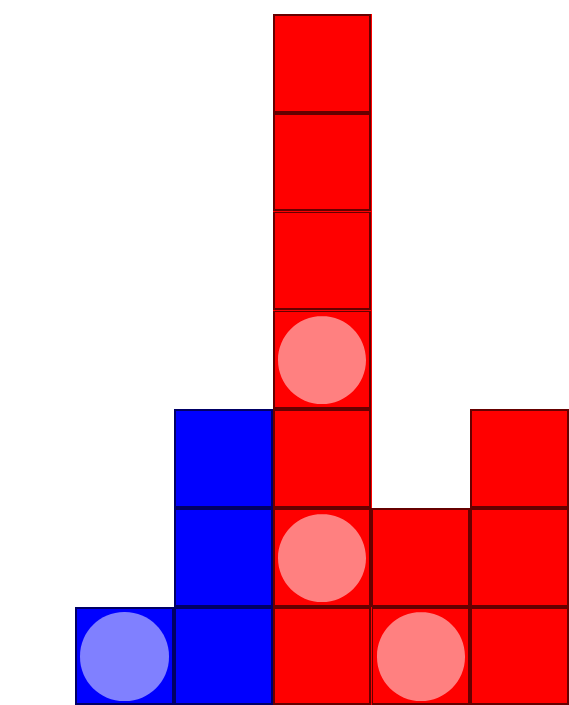}}
    \hfill
  \subfloat[One minus point.\label{fig:unfixablehole}]{%
        \includegraphics[width=0.3\linewidth]{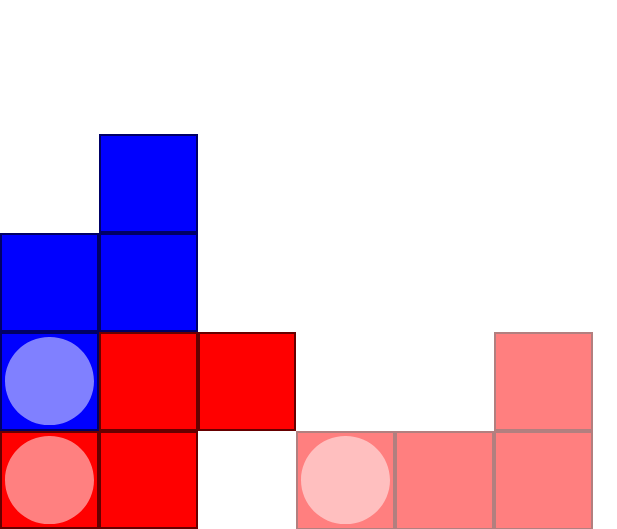}}
    \hfill
  \caption{Small examples to explain the game point system.}
  \label{fig:boardsit} 
\end{figure}

The goal of the game is to obtain the most points.
One point is awarded for every block, provided that it is connected to a group of at least three blocks. Not every block has to touch every other block in the group, as shown in Figure \ref{fig:groupthree}.

The \emph{I} block only touches the \emph{T} but not the \emph{L} on the far right. Since they together form a chained group of three, it counts as three points.
Blocks have to touch each other edge-to-edge. In Figure \ref{fig:nopoints}, the red player receives no points as the \emph{I} is only connected edge-to-edge to the blue \emph{L}.

A player loses one point per empty square (or hole) created, with a maximum of two minus points per turn.
Figure \ref{fig:unfixablehole} shows how one minus point for red would look like.
Moreover, the figure underlines a fundamental difference to video-game Tetris. In video-game Tetris, blocks slowly fall, and one could nudge the transparent \emph{L} under the \emph{S} to fill the hole by precise timing of an action. In Tetris Link, one can only throw pieces into the top and let them fall straight to the bottom. 
In the original rules, a dice is rolled to determine which block is placed. If a player is out of a specific block, then the player gets skipped. % Please do not use "he" if this can be avoided. Papers have been rejected for smaller reasons
Since every block is potentially one point, being skipped means missing out on one point. 

Due to the dice roll, Tetris Link is an imperfect information
game. The dice roll also causes random skips/point deductions for players. In this paper we omit the dice roll, analyzing the perfect information version of Tetris Link.
Note that we also focus on the two-player game only in this work, the three- and four-player versions are presumably even harder. However, our web based implementation for human test games\footnote{https://hizoul.github.io/contetro} can handle up to four players and can provide an impression of the Tetris Link gameplay.

\subsection{Verification that all games can end}
Each game of Tetris Link can be played to the end, in the sense that there are enough stones to fill the board. This is easy to see, by the following argument. 
The board is ten squares wide and twenty squares high so it can accommodate 200 individual squares.
Every player has twenty-five blocks, each consisting of four squares. There are always at least two players playing the game, so they are always able to fill the board. 

\begin{equation}
\begin{split}
playerPieces*squaresPerPiece*playerAmount \\
= 25*4*2 = 200
\end{split}
\end{equation}

\subsection{Branching Factor}
\label{sec:branch}
\begin{figure}
\centering
\includegraphics[width=0.47\textwidth]{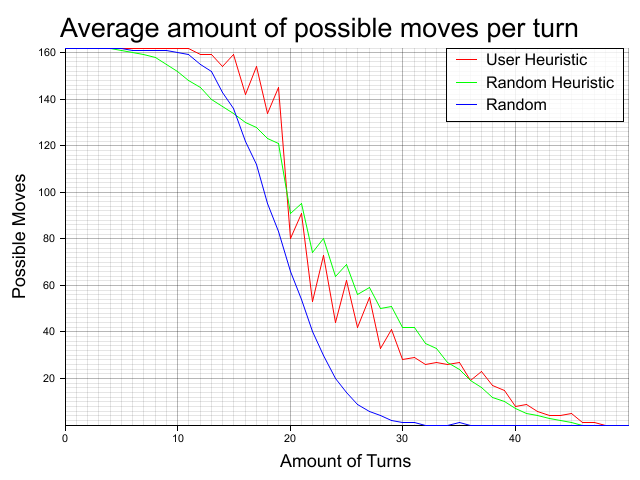}
\caption{The average number of possible moves throughout the game. See section \ref{sec:heuristic} for an explanation of how the heuristic agents work}
\label{fig:turnswithoutmargins}
\end{figure}

An essential property of a problem with respect to approaching it by means of a search algorithm is the branching factor. This is the number of possible moves a player can perform in one state\cite{edelkamp1998branching}.
In order to compute this number, we look at the number of orientations for each block.
The \emph{I} block has two orientations as it can be used either horizontally or vertically.
The \emph{O} block has only one orientation because it is a square.
The \emph{T} and the \emph{S} block have four different orientations for every side one can turn it to. The \emph{L} is an unusual case as it has eight orientations. Four for every side one can turn it to, but when one mirrors the shape, it has four additional sides to be turned to.
Hence, in total, nineteen different shapes can be placed by rotating or mirroring the available five base blocks.
Since the board is ten units wide, there are also ten different drop points per shape. In total, there can be up to 190 possible moves available in one turn.
However, the game rules state that all placed squares have to be within the bounds of the game board. Twenty-eight of these moves are always impossible because they would necessitate some squares to exceed the bounds either on the left or right side of the board.
Therefore, the exact number of maximum possible moves in one turn for Tetris Link is 162.
Since the board gets fuller throughout the game, not all moves are always possible, and the branching factor decreases towards the end.
In order to show this development throughout matches, we simulate 10,000 games. We depict the average number of moves per turn in Figure~\ref{fig:turnswithoutmargins}.

The first eight to ten turns, all moves are available regardless of the quality of play. After that, there is a slow but steady decline. Tetris Link is a game of skill: random moves perform badly. A game consisting of random moves ends after only thirty turns. Many holes with many minus points are created, and the game ends quickly with a low score.
The heuristic lines show that simple rules of thumb fill the board most of the time by taking more than forty turns. Furthermore, the branching factor in the midgame (turn 13-30) declines slower, and hence offer more variety to the outcomes.

We are now ready to calculate the approximate size of the state space of Tetris Link, in order to compare the complexity to other games. On average, across all three agents, a game takes 37 turns and allows for 74 actions per turn ($37^{74} \approx 1.45*10^{69}$).
The state-space complexity is larger than in Chess ($10^{47}$) but smaller than in Go ($10^{170}$).

\subsection{First move advantage}
\label{sec:first_advantage}
An important property of turn-based games is whether making the first move gives the player an advantage\cite{wikifirstmove}. To put the possible advantage into numbers, we let different strategies play against themselves 10,000 times to look at the win rate. The first six (\#1) or all (\#2) moves are recorded and checked for uniqueness.
\begin{table}[]
\centering
\begin{tabular}{|c|c|c|c|c|}
\hline
\textbf{Agent} & \textbf{Random} & \textbf{Random-H} & \textbf{User-H} & \textbf{Tuned-H} \\ \hline
\textbf{Win Rate \#1} & 47.84\%& 47.15\%& 71.93\%& 68.41\% \\ \hline
\textbf{Unique Games \#1} & 10,000& 2188 & 7& 29\\ \hline
\textbf{Win Rate \#2} & 48.16\% & 47\% & 71.65\%& 70\%\\ \hline
\textbf{Unique Games \#2} & 10,000& 10,000& 7& 50\\ \hline
\end{tabular}
\caption{First move advantage, over 10,000 games. The first six (\#1) or all turns (\#2) are compared for uniqueness to see whether the same games keep repeating. The -H in the agent name stands for Heuristic (see section \ref{sec:heuristic}).}
\label{fig:firstwin}
\end{table}

As can be seen in Table \ref{fig:firstwin}, the win rate for \textit{random heuristic} is almost 50\%. Although the win rate for the first player is higher for the \textit{tuned heuristics}, these numbers are not as representative because the heuristic repeats the same tactics over and over again resulting in only seven or twenty-nine unique game starts. If we repeat the same few games, then we will not know whether the first player has an advantage. Especially considering that at least until turn six, all moves are always possible, there are around $10^{13}$ or 18 Trillion\footnote{$Branching Factor ^{Turns} = 162^{6} = 18,075,490,334,784$} possible outcomes.
Since the \textit{random heuristic} has more deviation and plays properly as opposed to random moves, we believe that it is a good indicator of the actual first player advantage. Note that 47\% is close to an equal opportunity.
Different match history comparisons of Chess measure a difference of around two to five percent in win rate for the first player~\cite{wikifirstmove}. However, since neither Tetris Link nor Chess have been mathematically solved, one cannot be certain that there is a definite advantage.

\section{AI Player Design}
In this section, we describe the three different types of AI players that we implemented, based on heuristics, MCTS, and RL, respectively. 
%\subsection{Implementation}
%\label{sec:impl}
For the experiments (section \ref{sec:eval}), the game is coded in Rust and JavaScript (JS). The Rust version is written for faster experiments with MCTS and RL, and the JavaScript version is written to visually analyze games and also do a human play experiment.
Both implementations share a common game log format using JSON in order to enable interoperability.
To underline the importance of a performance-optimized version, we measured the average time it takes to simulate one single match where always the first possible move is made. The Rust implementation requires 590$\mu$s for that, whereas the JavaScript implementation needs 82ms.

\subsection{Heuristic}
\label{sec:heuristic}
We now describe the design of our heuristic player. A heuristic is a rule of thumb that works well most of the time~\cite{romanycia1985heuristic}. For Tetris Link, we identify four heuristic measures: 
% 
%So we extract four numerical values that we deem useful in decision making from the game board:
the number of connectable edges, the size of groups, the player score, and the number of blocked edges.
The number of blocked edges is the number of edges belonging to opponents that are blocked by the current players' blocks. All heuristic values are positively related to the chance of winning. 

Each parameter is multiplied by a weight, and the overall heuristic score is the sum of all four weighted values.
For every possible move in a given turn, the heuristic value is calculated, and the one with the highest value is chosen. If multiple moves have the same maximum value, a random one of these best moves is chosen.
The initial weights were manually set by letting the heuristic play against itself and detecting which combination would result in the most points gained for both players. We refer to this as \textit{user heuristic}.
We then use Optuna~\cite{akiba2019optuna}, a hyperparameter tuner, to tune a set of weights that reliably beat the \textit{user heuristic}. This version is called \textit{tuned heuristic}.

To achieve a greater variety in playstyle, we also test a \textit{random heuristic} which at every turn generates four new weights between zero to fifteen.
To have an estimate on the performance of the heuristic, we let actual human players (n=7) familiar with the game play against the heuristic via the JavaScript implementation. The \textit{random heuristic} achieved a win-rate of 23.07\% across 13 matches and the \textit{user heuristic} a win-rate of 33.33\% across six matches. The sample size is very small, but it still indicates that the heuristic is not particularly strong. This is supported by a qualitative analysis of the game played by the authors, based on our experience.
We conclude that our heuristic variants play the game in a meaningful way but are not particularly strong.

% 1. Heur
% 2. MCTS
% 3. RL
\subsection{MCTS}
For applications in which no efficient heuristic can be found, MCTS is often used, as it constructs a value function by averaging random roll-outs~\cite{browne2012survey}. 
Our MCTS implementation uses the standard UCT selection rule~\cite{kocsis2006bandit}. As further enhancements, we also use MCTS-RAVE~\cite{browne2012survey} and MCTS-PoolRAVE~\cite{rimmel2010biasing} to see whether the modifications help in improving the quality of results.
Furthermore, we experimented with improving the default (random) policy by replacing it with the heuristic. However, the heuristic calculation is so slow that it only manages to visit ten nodes per second.

MCTS is well-suited for parallelization, leading to more simulations per second and hence better play~\cite{browne2012survey}. We implemented tree parallelization, a frequently used parallelization~\cite{enzenberger2009lock}. In tree parallel MCTS, many threads expand the same game tree simultaneously. Using 12 threads, we visit 16258 nodes per second on average with a random default policy. To put this into perspective, this is $1.63e^{-9}\%$ of all $10^{13}$ possibilities in the first six turns. Thus, only a small part of the game tree is explored by MCTS, even with parallel MCTS.

%The next section provides evaluation results of MCTS, but first we will describe the design of the Reinforcement Learning environment.

% !!!!!!!!!!!!! here !!!!!!!!!!!!!!! aske
\subsection{Reinforcement Learning Environment and Agent}
\label{sec:envimpl}
A reinforcement learning environment requires an observation, actions and a reward~\cite{kaelbling1996reinforcement}, and an RL agent an algorithm as well as a network structure.
To prevent reinventing the wheel, we use existing code for RL, namely OpenAI gym~\cite{brockman2016openai} and the stable-baselines~\cite{stable-baselines}, which are written in Python. To connect Python to our Rust implementation, we compile a native shared library file and interact with it using Pythons \textit{ctypes}.
As RL Algorithm, we exclusively use the deep reinforcement learning algorithm PPO2\cite{schulman2017proximal}, without AlphaZero-like MCTS to further improve training samples. For the network structure, we increase the number of hidden layers from two layers of size 64 to three layers of size 128, because increasing the network size decreases the chances of getting stuck in local optima\cite{lawrence1997lessons}. We do not use a two-headed output, so the network only returns the action probabilities but not the certainty of winning like in AlphaZero\cite{silver2017mastering}.

The observation portrays the current state of the game field.
Inspired by AlphaGo which includes as much information as possible (even the "komi"\footnote{\textit{Komi} refers to the first turn advantage points~\cite{silver2016mastering}.}), we add additional information such as the number of pieces left per player, the players' current score and which moves are currently legal.
For the action space, we use a probability distribution over the possible moves. Probabilities of illegal moves are set to 0, so only valid moves are considered.
For the reward, we have three different options.
\begin{enumerate}
\item{Guided: $\frac{score+groupSize}{100}-scolding$}
\item{Score: $\frac{score}{100}$}
\item{Simple: $\pm1$ depending on win / loss}
\end{enumerate} The \textit{Guided} reward stands out because it is the only one that reduces the number of points via \textit{scolding}. If the chosen move was an illegal move, then the reward will be reduced, so the agent learns to only make valid moves. This technique is called reward shaping, and its results may vary~\cite{grzes2008plan}.

In order to detect which one of the three options is the most effective, we conduct an experiment.
%In order to figure out the effectiveness of the different reward functions (section \ref{sec:envimpl}), we reuse data collected for the reproducibility experiments (section \ref{sec:exprepro}). 
Per reward function, we collect the averages for the number of steps it took, the average reward achieved, and what the average score of the players was in the results.
Our results, shown in Table \ref{fig:res-reward}, indicate that the \textit{Guided} reward function works best. It only takes around 3183 steps on average to reach a local optimum, and the average scores achieved in the matches is the highest. The \textit{Score} reward function also lets the agent reach a local optimum, but it takes twice as long as the \textit{Guided} function, and the score is slightly lower as well.
The \textit{simple} reward function seems unfit for training. It never reached a local optimum in the 10,000 steps we allowed it to run and it got the lowest score in its games.
\begin{table}
\centering
\begin{tabular}{|c|c|c|c|}
\hline
\textbf{Reward Type} & \textbf{Steps} & \textbf{Episode Reward} & \textbf{Score} \\ \hline
\textbf{Guided} & 3183.49 & -0.17 & -5.6 \\ \hline
\textbf{Simple} & 10000.0 & -0.0 & -12.25 \\ \hline
\textbf{Score} & 6214.45 & -0.09 & -6.88 \\ \hline
\end{tabular}
\caption{Results of self-play with different reward types until either a local optimum or 10,000 steps have been reached. Step, Reward and Score show the average of all seeds.}
\label{fig:res-reward}
\end{table}

% Self-play is explained in RlAgents training setup
%{\bf 
%Now explain the RL player design:
%Plain RL, Deep RL, network architecture, two headed, or only value, self play against 5 opponents
%}

\section{Agent Training and Comparison}
\label{sec:eval}
For our experimental analysis, we first look at the performance of the MCTS agent (section \ref{sec:expmcts}) and the training process of the RL agents (section \ref{sec:exprl}).
Finally, we compare all previously introduced agents in a tournament to analyze their play quality and determine the currently best playing approach.

\subsection{MCTS Effectiveness}
\label{sec:expmcts}
\subsubsection{Setup}
\label{sec:setupmcts}
Initial test matches of MCTS against the \textit{user heuristic} resulted in a zero percent win rate, and a look at the game boards suggested near-random play.
We use a basic version of MCTS with random playouts because heuristic guidance was too slow.
AlphaZero has shown that even games with high branching factor such as Go can be played well by MCTS when guided by a neural network~\cite{silver2017mastering}.
However, without decision support from a learned model or a heuristic, we rely on simulations.
In order to see if this guidance is the reason for bad MCTS performance, we abuse the fact that the \textit{user heuristic} plays very predictably (section \ref{sec:first_advantage}).
We use the RAVE-MCTS variant (without the POOL addition), pre-fill the RAVE values with 100 games of the \textit{user heuristic} playing against itself and then let the MCTS play 100 matches against the \textit{user heuristic}. We repeat this three times and use the average value across all three runs.
We run this experiment with different RAVE $\beta$ parameter values. This parameter is responsible for the exploration/exploitation balancing and replaces the usual $C_p$ parameter. The closer the RAVE visits of a node reach $\beta$, the smaller the exploration component becomes.
Furthermore, we employ the slow heuristic default policy at every node in this experiment.
We simulate one match per step because otherwise, the one second thought time is not enough for the slow heuristic policy to finish the simulation step.

\subsubsection{Results}
\label{sec:evalmcts}
\begin{figure}
\centering
\includegraphics[width=0.45\textwidth]{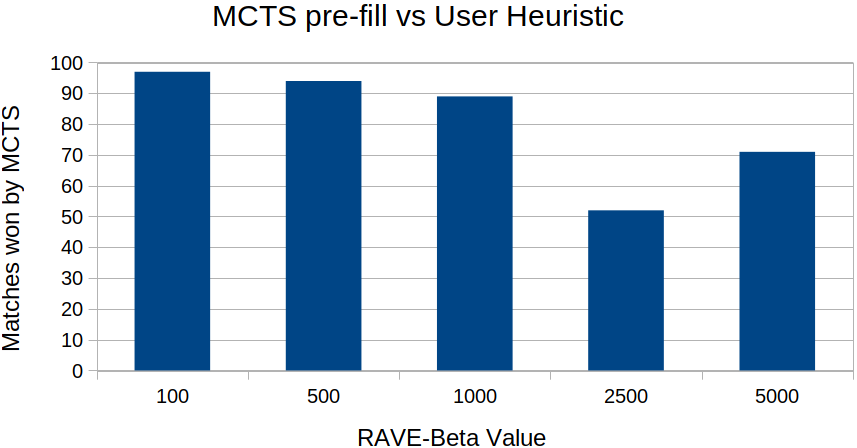}
\caption{Win rates of pre-filled MCTS playing against the \textit{User-Heuristic} compared by the RAVE-$\beta$ parameter.}
\label{fig:mctseval}
\end{figure}

Our MCTS implementation can play well with a decent win rate against the user-heuristic, as shown in Figure \ref{fig:mctseval}. 
This result underlines that in games with high branching factors, MCTS needs good guidance through the tree in order to perform well.
The declining win rate with a higher beta value suggests that exploration on an already partially explored game tree worsens the result if the opponent does not deviate from its paths. The rise in win rate for a $\beta$ value of 5000 after the large drop in 2500 underlines the effect that the randomness involved in the search process can have.

Even though the heuristic supported playout policy works well, we will still use a random playout policy for the tournament (section \ref{sec:exptournament}). Pre-filling the tree is very costly and would, therefore, provide an unfair advantage to the MCTS method.

%This is because there will be no pre-filling the tree in the tournament and with the heuristic default policy nearly no nodes will be visited of the empty tree.
We perform another small experiment in order to see how the branching factor influences MCTS performance: 
we run MCTS on different board sizes (2x2 to 11x11) of Hex against a shortest path heuristic. The result is striking: as long as the branching factor stays below 49 (7x7), MCTS wins up to 90\% of the matches. For larger branching factors, the win rate drops to 0\% quickly.

\begin{figure}
\centering
\includegraphics[width=0.5\textwidth]{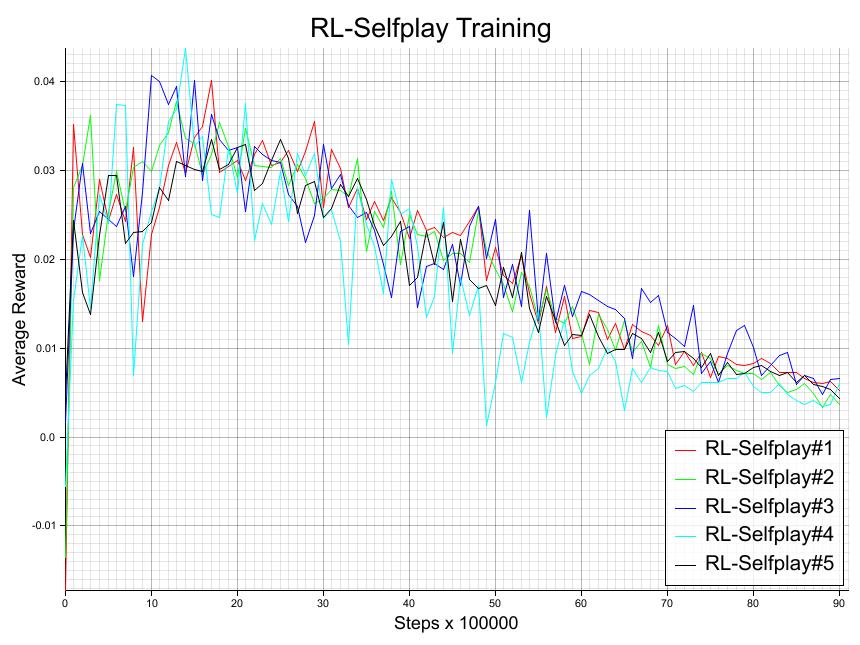}
\caption{The training process of \textit{RL-Selfplay} visualised by the average achieved reward per 100,000 Steps.}
\label{fig:train-self}
\end{figure}

\subsection{RL Agents Training}
\label{sec:exprl}
\subsubsection{Agents}
\label{sec:agents}
We define an RL agent as the combination of environment, algorithm and training opponent.
We use the guided reward function because it worked best in our experiment and call this agent \textit{RL-Selfplay}. (This is a neural network only RL, without MCTS to improve training samples.)

In addition to this rather simple agent, we introduce the \textit{RL-Selfplay-Heuristic} agent. It builds on a trained \textit{RL-Selfplay} agent where we continue training by playing against the heuristic. Observation and reward are the same as for \textit{RL-Selfplay}.

From the first turn advantage experiment, we know that the heuristic plays well even with random weights. That is why we also introduce an agent called \textit{RL-Heuristic}.
This agent gets the numerical observation as input and outputs four numbers that represent the heuristics weights (section \ref{sec:heuristic}). We use a modified version of the guided reward function:
\begin{equation}
\frac{(ownScore-opponentScore)+groupSize}{100}
\end{equation}

\begin{figure*}[t]
\centering
\includegraphics[width=1.0\textwidth]{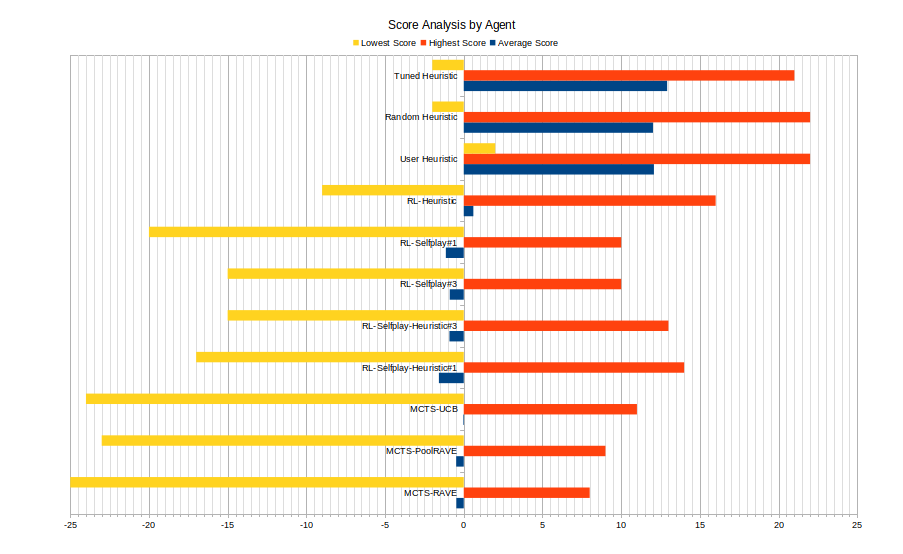}
\caption{Visualisation of the scores that agents achieved in the tournament. Agents are sorted by the skill rating in Fig.~\ref{fig:agent-skills}.}
\label{fig:agent-score}
\end{figure*}

Group size stands for the total number of stones that are connected with at least one other stone. This is added because we want the algorithm to draw a connection between the number of points gained and the number of connected stones.
However, mainly the difference in points between itself and the opponent is used as learning signal, so it aims for gaining more points than the opponent. Scolding is not necessary any more as we do not have to filter the output in any way.
\subsubsection{Setup}
\label{sec:setuprl}
In this section, we detail the training process of the RL agents.
Each training is done four times, and only the best run is shown.
Agents are trained with the default PPO2 hyperparameters, except for \textit{RL-Heuristic}, which uses hand-tuned parameters. 

Furthermore, we increase the hidden layer amount from two hidden layers with size 64 to three hidden layers with size 128 because increasing the network size decreases the chances of getting stuck in local optima\cite{lawrence1997lessons}.

When playing only against themselves, the networks still quickly reached a local optimum even with increased layer size. This optimum manifested in the same game being played on repeat and the reward per episode staying the same.
This repetition is a known problem in self-play and can be called "chasing cycles"~\cite{vinyals2019grandmaster}.
To prevent these local optima, we train five different agents against each other in random order.
To be able to train against other agents, we modified the \textit{stable-baselines} code.

\subsubsection{Results}
\label{sec:evalrl}

The training process for \textit{RL-Selfplay} is visualized in Figure \ref{fig:train-self}.
In the beginning, it keeps improving, but after peaking around 1.5 million steps, it only deteriorates. (Note that this is a form of Self-Play using the neural net only, without MCTS, as opposed to AlphaZero.)
Usually, a reward training graph should although jittery, steadily improve and climb in the reward achieved~\cite{mnih2015human}.

For \textit{RL-Selfplay-Heuristic} we use the two best candidates from \textit{RL-Selfplay}, namely \#3 after one million steps with a reward of 0.04, and \#1 after 1.5 million steps\footnote{The actual peak is at 1.7 million steps, but the model was only saved every 500,000 steps.} with a reward of 0.034.
The training of \textit{RL-Selfplay\#1-Heuristic} reaches its peak after 3.44 and \textit{RL-Selfplay\#3-Heuristic} after 3.64 Million steps with a reward of 0.032 and 0.024.
These are our first RL agents that can achieve a positive reward while playing against the heuristic.

The \textit{RL-Heuristic} training worked well, achieving mostly a positive reward.
But by looking at the output values, we realize the reward function design was unfortunate. It sets all weights to zero, except for the enemy block value which is fifteen and the number of open edges which varies between four and seven. So by negating the players score with the opponent's score, we have unwillingly forced the heuristic to focus on blocking the opponent over everything else. Needless to say with these weights, \textit{RL-Heuristic} rarely wins. Although it manages to keep the opponents score low, it does not focus on gaining points which leaves it with a point disadvantage.

\subsection{Tournament}
\label{sec:exptournament}
\subsubsection{Setup}
\label{sec:setuptournament}
In the tournament, we will pit all previously shown AI approaches against each other.
Every bot will play 100 matches against every other bot. We have five different RL bots, three MCTS bots and three heuristic bots.
The bots skill will be compared via a Bayesian Bradley Terry (BBT) skill rating\cite{weng2011bayesian}. The original BBT\cite{weng2011bayesian} uses a skill rating in the range of 0 to 50, similar to TrueSkill\cite{herbrich2007trueskill}. By changing the $\beta$ parameter of the rating function, we change the range from 0 to 3000, so it is similar to the standard ELO range\cite{glickman1999rating}. 

\subsubsection{Results}
\label{sec:evaltournament}
\begin{figure}
\centering
\includegraphics[width=.5\textwidth]{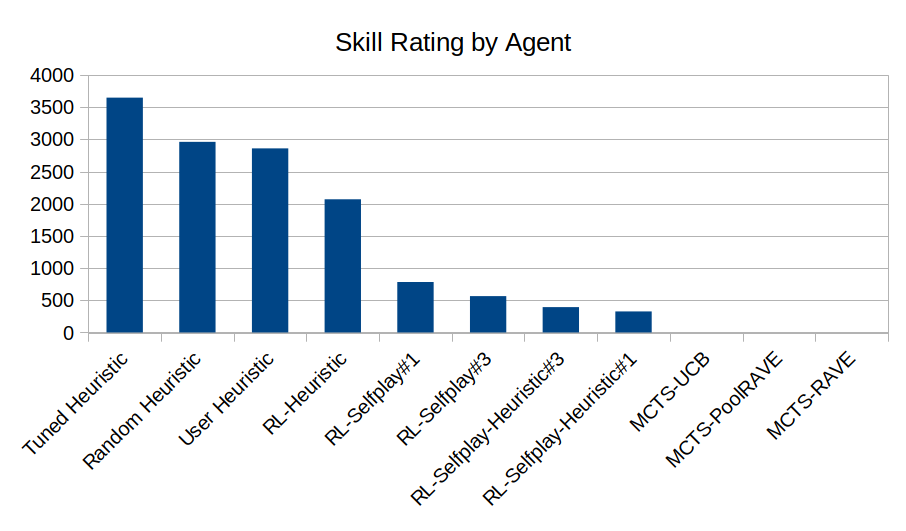}
\caption{The skill rating of the agents that participated in the tournament.}
\label{fig:agent-skills}
\end{figure}
The final skill rating is portrayed in Figure \ref{fig:agent-skills}.
The three heuristic agents take the top 3, followed by RL and MCTS. Remarkably, the \textit{tuned heuristic} performed best, even though it is only optimized to play well against the \textit{user heuristic}, but yet it performs best across all agents.

Seeing \textit{RL-Heuristic} as the best \textit{RL} approach shows that the other RL agents are far from playing well.
Yet all RL agents consistently beating MCTS with random playouts proves that the agents definitely learned to play reasonable.

It is interesting to see that the MCTS-UCB (14\% win rate) variant performed best because the other two variants [RAVE (0.02\%), PoolRAVE (0.04\%)] were conceived in order to improve the performance of UCB via slight modifications\cite{rimmel2010biasing}.

The skill rating omits information about the quality of the individual moves. To gain further insight into that, we provide Figure \ref{fig:agent-score}. Here, we can see that every agent manages at least once to gain 8 points or more. This means that every agent had at least one match it played well.
Looking at the lowest achieved scores and average scores, we find that every agent except for the pure heuristic ones plays badly, considering that on average, they only make $\pm3$ points.

\section{Conclusion and Future Work}
% mention bad reward
% vanishing gradient => LSTM?
Board game strategy analysis has been done for decades, and especially games like Chess and Go have seen countless papers analyzing the game, patterns and more to find the best play strategies~\cite{silver2017mastering}.
We contributed to that field by taking a close look at the board game Tetris Link.
While the strategy is key to winning, some games, such as Hex, give the first player a definite advantage. We have experimentally shown that there is no clear advantage for the starting player in Tetris Link (section \ref{sec:first_advantage}). 

We have implemented three game playing programs, based on common approaches in AI: heuristic search, MCTS, and reinforcement learning. Despite some effort, none of our programs was able to beat human players.

In doing so, we have obtained an understanding of why it may be hard to design a good AI for Tetris Link:
\begin{itemize}
    \item Especially at the beginning, the branching factor is large, staying at around 160 for at least the first six turns.
    \item Many moves cannot be reversed. The unforgivingness for these moves may make it  harder to come up with a decent strategy, as generally postulated by~\cite{CookR19}.
    \item Many rewards in the game stack --- they  come  delayed after multiple appropriate moves because  groups of pieces count and not single pieces.
\end{itemize}

All this holds true for the simplified version we treat here: no dice, only two players. Adding up to two more players and dice will also make the game harder.

With a solid understanding of the game itself, we investigated different approaches for AI agents to play the game, namely heuristic, RL and MCTS.
We have shown that all tested approaches can perform well against certain opponents.
The best currently known algorithmic approach is the tuned heuristic, although it can not consistently beat human players.

Training an RL agent (section \ref{sec:exprl}) for Tetris Link has proven to be complicated. Just getting the network to produce positive rewards required much trial and error, and in the end, the agent did not perform well even when consistently achieving a positive reward.
We believe the learning difficulty in Tetris Link comes from the many opportunities to make minus points in the game. One turn offers at most one plus point, or three and more if a group is connected, but that means that the previous two or more turns at most gave zero points if not even more minus points. Hence recovering from minus points is difficult, meaning small mistakes have graver consequences.\footnote{Note that the RL agent did not use MCTS-based self-play as  AlphaZero \cite{silver2017mastering}, but  a neural network, as used for Atari \cite{mnih2015human}.} 

%Reproducible training is also a problem we stumbled over. It is well known that neural network algorithms results can differ strongly by just changing little of the input (e.g. data, or parameters such as the random seed). Nevertheless, if the input stays the same, then the results should also stay unchanged. However, in our experiments, that was not the case, and less than half of the time, training was actually reproducible. This is a known problem related to \textit{stable-baselines} and \textit{TensorFlow}, of which developers are aware and actively working on~\cite{nvidiatensor,stablerepro}.
%Our reproducibility problems only focused on the code that runs it, but there are other reproducibility problems in the AI field. At the ICLR challenge in 2018, where they asked researchers to try to reproduce papers, only 32.7\% of checked papers were mostly reproducible~\cite{reprotalk}\footnote{To enable reproducibility of this work all used code is available at https://github.com/Hizoul/masterthesis}.
%Luckily people are working on identifying sources of reproducibility issues and propose solutions\cite{henderson2018deep}. Software like OpenAI Gym tackles one of the problems. Namely, uniform environments for more reliable benchmarks across different RL algorithm implementations. So with our Tetris Link Gym implementation, we offer researchers another environment for benchmarking.
%The used software environment is also still rapidly growing and developing, as can be seen with OpenAI's standardisation of \textit{PyTorch}\cite{openaitorch}.

Although MCTS performed poorly in our tournament, we have shown that with proper guidance through the tree MCTS can perform nicely in Tetris Link and Hex (section \ref{sec:expmcts}). That is why a combination where RL guides an MCTS through the tree might work well, e.g. AlphaZero~\cite{silver2017mastering} or MoHex v3~\cite{gao2017move}, and is something to try in future work.

We invite the research community to use  our code and improve upon our approaches\footnote{https://github.com/Hizoul/masterthesis}. 

% Future: try smaller boards, try zero learning
%\newpage
\bibliographystyle{IEEETran}
\bibliography{report}

\end{document}